\documentclass{article}

\usepackage[dblblindworkshop, final]{neurips_2026}

\usepackage[utf8]{inputenc} 
\usepackage[T1]{fontenc}    
\usepackage{hyperref}       
\usepackage{url}            
\usepackage{xurl}           
\usepackage{booktabs}       
\usepackage{graphicx}       
\usepackage{amsmath}        
\usepackage{amsfonts}       
\usepackage{nicefrac}       
\usepackage{microtype}      
\usepackage{xcolor}         
\usepackage{listings}       
\usepackage[most]{tcolorbox} 

\newtcblisting{promptbox}[1]{
  listing only, breakable, enhanced jigsaw,
  colback=black!3, colframe=black!50, boxrule=0.4pt, arc=1pt,
  left=4pt, right=4pt, top=3pt, bottom=3pt,
  title={#1}, fonttitle=\footnotesize\bfseries,
  listing options={basicstyle=\ttfamily\scriptsize, breaklines=true,
    columns=fullflexible, keepspaces=true, breakindent=0pt,
    literate={ł}{{\l{}}}1}}

\title{RLPE: Teaching LLMs to Prompt with \\Reinforcement Learning}

\author{Rahul Singhal\\
\texttt{rahul@distyl.ai}
\And
Karime Maamari\\
\texttt{karime@distyl.ai}
\And
Pradyumna Tambwekar\\
\texttt{pradyumna.tambwekar@distyl.ai}
}

\begin{document}

\maketitle

\begin{abstract}
  Prompting is a widely used method for adapting general-purpose LLMs to
  domain-specific tasks. Effective prompts can be tedious to create manually, so prompt
  writing is increasingly handed to LLMs, with recent work training
  LLMs to write prompts for other models. These techniques either
  write prompts without observing
  how the task model behaves, blind to the
  failure modes the prompt should address, or must run alongside the
  task LLM at inference time, adding overhead to every query.
  Neither reflects how prompt engineering is often practiced. To close
  this gap, we introduce Reinforcement Learned Prompt Engineering
  (\textsc{RLPE}), which trains a model to analyze traces of a separate
  task model and write a prompt targeting the failures it observes.
  \textsc{RLPE} rewards the trained model strictly on prompt
  performance, so it learns the prompting strategies that improve the
  task model. The prompt is generated once and reused across the task, so
  inference does not require the trained model. Since \textsc{RLPE}
  needs only query access to the model it prompts, it is also
  compatible with closed-source models. We apply \textsc{RLPE} to Enigmata, BIG-Bench Extra Hard, and PlanBench. We train a 9B-parameter model
  that writes better prompts than state-of-the-art frontier models
  (\texttt{gpt-5.6-sol} and \texttt{claude-opus-5}) and prompt optimizers (GEPA and TextGrad)
  on 2/3 benchmarks. Although the model is never explicitly trained to iterate
  on prompts, this capability emerges at a level that matches or beats
  frontier models. The learned prompting capability also generalizes, with the trained model
  improving task models 6x larger than the one it trained with,
  including ones from entirely different model families, and matching
  \texttt{claude-opus-5} on a fourth benchmark held out from training.

\end{abstract}


\section{Introduction}
\label{sec:intro}

General-purpose large language models (LLMs) are adapted to specific tasks largely through prompting \citep{liuPretrainPromptPredict2023}, so finding effective prompts grows more valuable as LLMs take on more tasks. Automated prompt-optimization tools use LLMs
to generate and search for high-performing prompts and have shown broad success \citep{agrawalGEPAReflectivePrompt2025,
yuksekgonulTextGradAutomaticDifferentiation2024}. But these tools are bounded by the prompting ability of the models
powering them, which remains largely understudied.

Recent work has begun directly training models to write prompts.
PRewrite \citep{kong_prewrite_2024}, PRL \citep{batorski_prl_2025},
and Prompt-MII \citep{xiao_prompt-mii_2025} train models to produce
prompts from task descriptions or labeled examples. Trained this way,
a model never observes the prompted model's behavior,
making it unclear what its prompts should correct. Advisor Models \citep{asawa_how_2025} and
Prompt-R1 \citep{liu_prompt-r1_2025} train a model that works
alongside the prompted model, advising it on each query. Here the
trained model does observe the prompted model's behavior, but it must
run with the prompted model on every query, adding inference overhead
and limiting practical use. Neither captures the
process of prompt engineering: inspecting outputs, spotting failure
modes, and incorporating the fixes into a reusable prompt.


We introduce Reinforcement Learned Prompt Engineering (\textsc{RLPE}),
which trains a prompt-writing model under conditions that better
mirror prompt engineering practice. We train an \textit{advisor model}
to write prompts for a \textit{task model}, the model performing the
task. To write a prompt, the advisor is shown a batch of
traces, each with a question the task model attempted, its
response, and the response's correctness, exposing precisely how the task
model performs. The advisor generates a static prompt used for every question the
task model answers, so the advisor never runs per query. During training, we run the
task model with each prompt and reward the advisor by the
measured performance. \textsc{RLPE} is practical to use. It needs only query access to
the task model, so it is compatible with closed-source models
(Section~\ref{sec:results-generalization}). Also, \textsc{RLPE} training is
cheap, requiring modest compute and no reference prompts or LLM
judges, since performance alone provides the reward.

\begin{figure}[t]
  \centering
  \includegraphics[width=\textwidth]{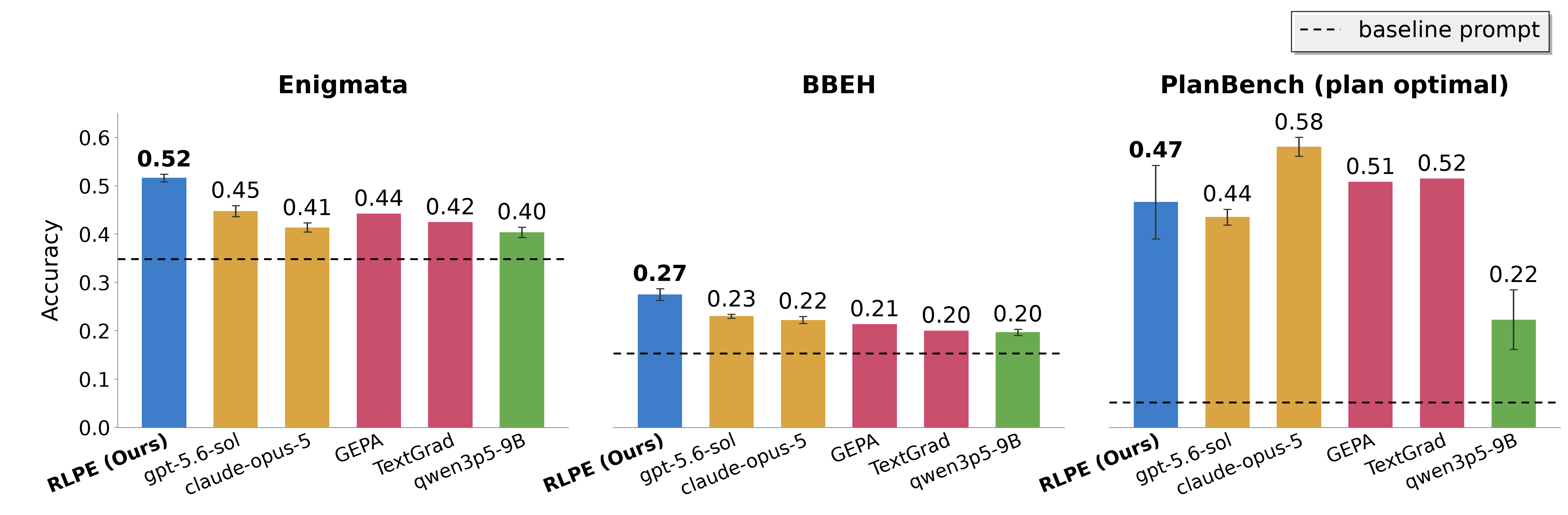}
  \caption{\textbf{\textsc{RLPE} writes the strongest prompts on 2/3
  benchmarks.} Bars = prompt performance (SE bars if
  performance is averaged); dashed line = baseline prompt performance.
  \textsc{RLPE} (bolded) leads on Enigmata (0.52 \textsc{RLPE} vs.\ 0.45
  \texttt{gpt-5.6-sol}) and BBEH (0.27 \textsc{RLPE} vs.\
  0.23 \texttt{gpt-5.6-sol}), and on PlanBench is competitive
  with every method except \texttt{claude-opus-5} (0.47 \textsc{RLPE} vs.\ 0.58 \texttt{claude-opus-5}).
  One untrained-policy generation on PlanBench failed to parse (scored
  0).}
  \label{fig:main-results}
\end{figure}

We train a 9B-parameter model via Group Relative Policy
Optimization (GRPO) \citep{shaoDeepSeekMathPushingLimits2024} that writes
more performant prompts than state-of-the-art frontier models and prompt optimizers. It surpasses \texttt{gpt-5.6-sol}, \texttt{claude-opus-5}, GEPA, and TextGrad on Enigmata
\citep{chen_enigmata_2025} and BIG-Bench Extra Hard (BBEH)
\citep{kazemi_big-bench_2025}, and matches nearly all baselines on
PlanBench \citep{valmeekam_planbench_2022}
(Figure~\ref{fig:main-results}). These prompts are written from scratch, using
only traces with no existing prompt to build on. Practitioners more often revise prompts than write new ones \citep{tafreshipourPromptingWildEmpirical2025}, so we evaluate iteration as a skill, in which the advisor is given a prompt to
revise as well.
Although prompt iteration is never directly trained, the
capability emerges, with \textsc{RLPE} matching or beating frontier
models at revising prompts (Figure~\ref{fig:fixed-start}). 

Training for effective prompting is far less useful without
transfer beyond the task model and dataset used in training. We test the advisor along these generalization axes. Although trained with one task model, our advisor improves task
models six times larger and models from other model
families (Figure~\ref{fig:taskmodel-gen}). On JustLogic
\citep{chen_justlogic_2025}, a fourth, held-out benchmark,
the trained model consistently beats the untrained policy and is competitive
with frontier models (Table~\ref{tab:justlogic}). These results suggest the
learned prompting ability is general, not narrowly tuned to its
training setup.

An overview of our contributions is as follows:
\begin{itemize}
\item We introduce Reinforcement Learned Prompt Engineering
  (\textsc{RLPE}), a training framework that mirrors prompt
  engineering more faithfully than previous work: an advisor model
  reads task-model traces, writes a prompt, and is rewarded by how
  well that prompt performs.
\item We train a 9B-parameter advisor model that beats state-of-the-art
  frontier models (\texttt{gpt-5.6-sol} and \texttt{claude-opus-5}) and
  prompt-optimizer tools (GEPA and TextGrad) on two out of three benchmarks. This advisor also effectively iterates on prompts, a capability
  that emerges without direct training and matches or exceeds
  frontier models.
\item We provide evidence that the learned prompting capability generalizes. The trained advisor
  improves task models six times larger than its training task model, as
  well as models from entirely different model families.
  Additionally, all trained advisors beat the untrained policy and match \texttt{claude-opus-5} on a held-out
  benchmark not seen during training.
\end{itemize}

\section{Related Work}
\label{sec:related}

\paragraph{RL for LLMs.}
Reinforcement learning is now a widely used post-training method for
improving targeted capabilities such as mathematical reasoning,
coding, and agentic tool use \citep{qianToolRLRewardAll2025}. Beginning with reinforcement learning
from human feedback (RLHF), the field trained models toward human
preferences \citep{ouyangTrainingLanguageModels2022}. More recently,
focus has shifted to reinforcement learning with
verifiable rewards (RLVR), which samples many rollouts per prompt and
updates the model toward the highly rewarded ones
\citep{deepseek-ai_deepseek-r1_2025}. We apply the same technique, but instead of improving a model's own
task performance, we train it to write prompts that improve the
performance of a separate task model we may only be able to query.

\paragraph{Prompt optimization.}
Prompt-optimization tools automate the search for a prompt that
maximizes task performance. Early methods such as APE
\citep{zhouLargeLanguageModels2023} and OPRO
\citep{yangLargeLanguageModels2024} treat the LLM entirely as a
prompt generator, sampling instructions and keeping the highest-scoring ones.
More recent tools instead have the LLM read the task model's failures
and revise prompts: GEPA
\citep{agrawalGEPAReflectivePrompt2025} uses an evolutionary approach
to refine prompts, TextGrad
\citep{yuksekgonulTextGradAutomaticDifferentiation2024} performs
gradient-descent-style optimization with natural-language feedback,
and PrefPO \citep{singhal_prefpo_2026} drives revisions with pairwise
preferences over outputs. These tools are only as good as the LLM behind them, which is particularly problematic
because prompting ability is not often evaluated directly. We respond by
measuring how well models prompt across different settings and
training a model to improve that capability.

\paragraph{RL for prompting.}
A growing line of work trains prompting models
\citep{dengRLPromptOptimizingDiscrete2022, kong_prewrite_2024}.
Some approaches train a model that
steers the task model one query at a time: Advisor Models
\citep{asawa_how_2025} appends advice to each question, Prompt-R1
\citep{liu_prompt-r1_2025} runs a multi-turn agent that prompts the
task model and answers itself, and others write per-instance critiques
or reflections \citep{xie_teaching_2025, yao_retroformer_2023}.
These approaches must run the trained model alongside the task model for every query, making it harder to scale. Other approaches avoid
this per-query overhead by training a model to write one shared
prompt for the task. PRL \citep{batorski_prl_2025} trains a prompt
generator that writes prompts from just task descriptions, and
Prompt-MII \citep{xiao_prompt-mii_2025} learns to generate
instructions from labeled examples across many datasets. These approaches do not observe
the task model's behavior, hiding what the prompt must fix. Our advisor addresses both shortcomings of prior work. Its
output is a standalone prompt, so the advisor is unnecessary at
inference time, and its input is the task model's traces, exposing remaining
failure modes.


\section{\textsc{RLPE}: Reinforcement Learned Prompt Engineering}
\label{sec:approach}

\subsection{Problem Statement}
\label{sec:approach-setup}


\paragraph{Preliminaries.} We train an \textit{advisor model},
$\pi_{\mathrm{adv}}$, a policy that writes prompts for a separate task
model based on observed traces of its behavior. The \textit{task
model}, $\pi_{M}$, is a fixed language model we wish to prompt
effectively; we assume only query access to it. We define a \textit{prompt} $p$ as a system prompt and a user
prompt template into which a given question $q$ is inserted. We
write $p(q)$ for the prompt with question $q$ in place and sample
responses $y \sim \pi_{M}(\cdot \mid p(q))$ (full example prompts in
Appendix~\ref{app:prompt-example}). Throughout our experiments, we use a baseline prompt, which keeps the
system prompt empty and passes the question through unchanged as the
user prompt, adding brief answer-formatting instructions when a
question does not specify them (the exception is PlanBench, whose
baseline is the benchmark's official query containing the rules of the
world and a one-shot example). A task is a pair
$(\mathcal{D}, R)$, a distribution over questions and a
programmatic verifier $R(q, y) \in \{0, 1\}$ that scores a response. A
\textit{trace} $\tau = (q, y, r)$ records a question, the task model's
response to it under some prompt, and the score $r = R(q, y)$; we
write $T = \{\tau_1, \dots, \tau_k\}$ for a batch of $k$ traces.

\label{sec:approach-tasks}
\paragraph{Two prompt engineering settings.} We divide prompt engineering into two main settings, corresponding to two ways of conditioning
$\pi_{\mathrm{adv}}$: \textit{from-scratch} and \textit{iterative} generation.

For from-scratch generation, the advisor is shown
only a batch of traces $T$ produced under the baseline prompt and must write a
new prompt for the benchmark, inferring from the traces what the task model
gets wrong and what instructions would fix it:
\[
  p \;\sim\; \pi_{\mathrm{adv}}(\cdot \mid T).
\]
For iterative generation, the advisor is additionally shown the prompt
$p$ currently in use, together with the corresponding traces $T$ produced by running
$p$, and must write a revised prompt that addresses the issues $p$ has:
\[
  p' \;\sim\; \pi_{\mathrm{adv}}(\cdot \mid p, T).
\]
Separating prompt engineering into these two settings aligns our
setup with real practice, where practitioners both generate new
prompts and iterate on existing ones. 


\paragraph{Objective.} Both settings share a single goal: produce a prompt $p$
that performs as well as possible on the benchmark, that is, that maximizes
\[
  \mathrm{Acc}(p) \;=\; \mathbb{E}_{q \sim \mathcal{D}}\,
  \mathbb{E}_{y \sim \pi_{M}(\cdot \mid p(q))}\big[R(q, y)\big],
\]
the expected accuracy of $\pi_{M}$ under $p$ over the task's full question
distribution $\mathcal{D}$. We study
from-scratch and iterative generation as two methods of pursuing this same
objective.

\subsection{Training Procedure}
\label{sec:approach-training}

We train $\pi_{\mathrm{adv}}$ with GRPO on the from-scratch setting only. Each
emitted prompt $p$ is rewarded by
\[
  \mathrm{Acc}_T(p) \;=\; \frac{1}{|T|}\sum_{(q,y,r) \in T}
  \mathbb{E}_{y' \sim \pi_{M}(\cdot \mid p(q))}\big[R(q, y')\big],
\]
the accuracy over the shown traces $T$, since this is what we can feasibly
compute during training. If the advisor's output is misformatted, it is penalized, and the task model is never run.

We deliberately train only on the from-scratch setting. We are
particularly interested in whether the trained advisor can iterate on
prompts without explicit training, to better understand if our training
procedure can generalize to unseen prompt engineering settings. We
treat joint training as a natural extension for future work.


\section{Experimental Setup}
\label{sec:method}

\subsection{Data}
\label{sec:method-data}

We use three benchmarks across model capabilities for training: BBEH
(multi-step reasoning problems), Enigmata (various self-contained puzzles), and PlanBench (block-stacking plan-generation
questions). We additionally evaluate on JustLogic (deductive logic reasoning problems) as a
held-out benchmark never seen during training. We selected benchmarks
against two criteria. First, each benchmark must be reasoning-sensitive, showing large accuracy gains when increasing
\texttt{gpt-oss-20b}'s reasoning effort from \texttt{low}, since we believe most of a
prompt's effect is guiding reasoning
(full details in Appendix~\ref{app:effort-curve}). Second, each benchmark must provide
a programmatic verifier, which keeps grading cheap and fast for the
many prompt evaluations training requires. 

We split BBEH's original data into train, validation, and test sets.
For Enigmata and PlanBench we add synthetic examples from the
benchmarks' official generators\footnote{\url{https://github.com/BytedTsinghua-SIA/Enigmata} and \url{https://github.com/karthikv792/LLMs-Planning}.} to the original
data. This extra data
is needed because our prompt-sensitivity filtering, applied only to
the training data, requires more raw questions than the benchmarks
provide. For PlanBench, we train on its official
plan-generation task but evaluate on its official optimal-planning
variant, defined over the same questions but additionally requiring plans to be as
short as possible, which resolves an ambiguity in the original task
(Appendix~\ref{app:planbench-t2}). For JustLogic, which serves only as a
held-out generalization test, we draw traces from the benchmark's
official train split and score on a stratified sample of its official
validation split. Exact splits for every benchmark are listed in
Appendix~\ref{app:splits}.

\paragraph{Prompt-sensitivity filtering.}
Many benchmark questions are insensitive to prompting, meaning the task
model answers them correctly, or fails them, regardless of the prompt
provided. These questions provide no training signal, so we filter them out of
the training data, leaving the validation and test splits untouched. To identify them, we ran high-performing and low-performing prompts
over the data and kept only the questions answered correctly more often
under the high-performing prompts. For some benchmarks this cut the
training data by up to 50\% (full details in
Appendix~\ref{app:sensitivity}).

\subsection{Models}
\label{sec:method-models}

\paragraph{Advisor models.}
We train \texttt{qwen-3.5-9b} \citep{qwenteamQwen35Native2026}, a strong
open-weight model with far fewer parameters than both the
frontier models and the task model it writes prompts for. We compare against several other models that we use to write prompts
without training, including frontier models \texttt{gpt-5.6-sol}
\citep{openaiGPT56SystemCard2026} and \texttt{claude-opus-5}
\citep{anthropicClaudeOpus5System2026}, and the untrained \texttt{qwen-3.5-9b} to
isolate the effect of training.

\paragraph{Task models.}
The primary task model, used for training and our main results,
is \texttt{gpt-oss-20b} \citep{openaiGptoss120bGptoss20b2025}
(\texttt{low} reasoning effort, temperature~1.0). To test
generalization across task models, we additionally evaluate trained prompts
on \texttt{gemma-4-26b} \citep{gemmateamGemma4Technical2026},
\texttt{gpt-oss-120b} \citep{openaiGptoss120bGptoss20b2025},
\texttt{gpt-5-nano}
\citep{openaiGPT5SystemCard2026}, and \texttt{gpt-5.4-nano}
\citep{openaiIntroducingGPT54Mini2026}, all at \texttt{low} reasoning
effort, and \texttt{claude-haiku-4-5}
\citep{anthropicClaudeHaiku45System2025} with a 32{,}000-token
thinking budget.

\subsection{Training Setup}
\label{sec:method-rl}

\paragraph{Training configuration.}
We train a separate advisor per benchmark with GRPO (LoRA rank~32, learning rate
$3\times10^{-5}$, KL penalty~0) for 50 steps with 32 groups per step and 8
rollouts per group, totaling 12{,}800 rollouts. The trained model generates up to 32{,}768 tokens at
temperature~1.0 per rollout. Each rollout shows the advisor $k=10$ traces and instructs it to create a new
prompt (full instructions in Appendix~\ref{app:metaprompt}). The traces are generated with the baseline prompt. We settled on ten
traces as an intentional tradeoff: enough for a reliable read on prompt
performance, but few enough that the method stays usable in
low-data settings. Training and inference ran on NVIDIA B200 GPUs (full compute details in Appendix~\ref{app:compute}).


\paragraph{Reward.}
Rollouts are rewarded by $\mathrm{Acc}_T(p)$, as defined in
Section~\ref{sec:approach-training}; a misformatted output from the trained policy receives
a fixed reward of $-0.1$.

\subsection{Baseline Configurations}
\label{sec:method-baselines}

\textbf{Frontier models.} \texttt{gpt-5.6-sol} and
  \texttt{claude-opus-5} are passed the same instructions used during training, at \texttt{medium} reasoning effort and default temperature, with a
  32{,}768-token output cap.
  
\textbf{Untrained policy.} We run
    the \texttt{qwen-3.5-9b} base model with the same instructions used during training, at \texttt{medium} reasoning effort and temperature~1.0.
    
\textbf{GEPA.} GEPA is an evolutionary prompt optimizer that leverages an LLM to read
  the task model's failures and propose a revision
  \citep{agrawalGEPAReflectivePrompt2025}. We run it at its highest
  optimization budget (\texttt{auto="heavy"}) with default parameters
  otherwise.

\textbf{TextGrad.} TextGrad casts prompt optimization as gradient
  descent in natural language, using LLMs to produce textual feedback
  on the current prompt's failures and apply it as
  an update \citep{yuksekgonulTextGradAutomaticDifferentiation2024}. We run with default parameters.

Both GEPA and TextGrad use \texttt{gpt-5.6-sol} at \texttt{medium}
reasoning effort with a 32{,}768-token cap as the LLM powering the
optimization. The final prompt for each is selected as the
highest-scoring candidate on the full validation set (full details in
Appendix~\ref{app:optimizer-splits}). This feedback is unavailable to the advisor models, making the optimizers particularly
well-resourced baselines.

\subsection{Evaluation}
\label{sec:method-eval}

\paragraph{Prompt optimizers.}
Each optimizer's best-performing prompt is
scored on the test set.

\paragraph{Advisor models.}
For each benchmark, we draw $B=10$ batches each with $k=10$ traces from
the validation set. In the from-scratch setting, each advisor model
receives one batch at a time with the same instructions used
during training, and writes one prompt per batch (full from-scratch instructions in Appendix~\ref{app:metaprompt}). We score the
$B$ prompts on the test set and report their mean accuracy. In the
iterative setting, each advisor model additionally receives an initial
prompt to revise in its instructions, together with the traces the task
model produced under that prompt (full iteration instructions in Appendix~\ref{app:iter-metaprompt}). We again report the mean test-set
accuracy of the $B$ revised prompts. All reported error bars are one
standard error of the mean, computed across the $B$ prompt scores.


\section{Results}
\label{sec:results}

We train three advisor models with \textsc{RLPE}, one each on
Enigmata, BBEH, and PlanBench. The following sections evaluate them in the from-scratch and iterative settings and test generalization.

\begin{figure}[t]
  \centering
  \includegraphics[width=0.8\textwidth]{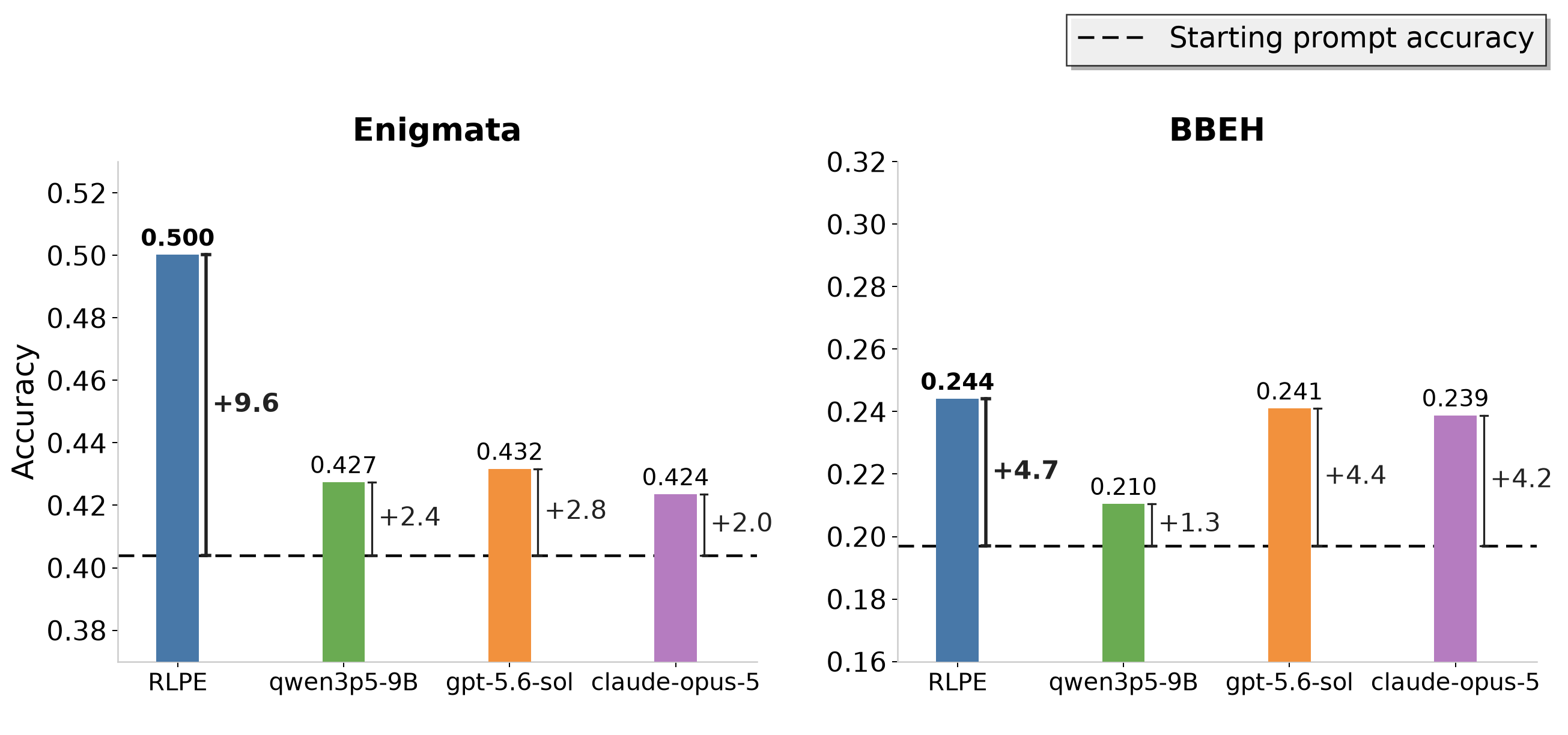}
  \caption{\textbf{\textsc{RLPE} matches or beats frontier models at
  iterating on prompts.} Every advisor model (trained, untrained, and
  frontier) revises the same ten starting prompts, taken from the
  untrained policy's from-scratch prompts in
  Figure~\ref{fig:main-results}. Bars = mean score of the ten
  revised prompts; dashed line = starting prompts' score; brackets
  = performance gains. On Enigmata, \textsc{RLPE} (bolded) gains more
  than triple any other advisor model (+9.6 vs.\ +2.8
  \texttt{gpt-5.6-sol}), and on BBEH, \textsc{RLPE} (+4.7) is on par
  with the frontier models (+4.4 \texttt{gpt-5.6-sol} and +4.2
  \texttt{claude-opus-5}).}
  \label{fig:fixed-start}
\end{figure}

\subsection{From-Scratch Setting}
\label{sec:results-scratch}

\paragraph{Setup.}
We compare the trained advisor against the untrained
\texttt{qwen-3.5-9b} policy, the frontier models, and the prompt
optimizers in the from-scratch setting across all three datasets, following the protocol in Section~\ref{sec:method-eval},
with the baseline prompt as a reference point.

\paragraph{Results.}
Figure~\ref{fig:main-results} shows \textsc{RLPE} more than doubles
the untrained policy's gains over the baseline prompt on every
benchmark (16.8 vs.\ 5.5 points on Enigmata, 12.2 vs.\ 4.4 on BBEH,
and 41.4 vs.\ 17.1 on PlanBench). \textsc{RLPE} beats frontier models on Enigmata (0.516 \textsc{RLPE} vs.\
0.448 \texttt{gpt-5.6-sol} and 0.414 \texttt{claude-opus-5}) and BBEH
(0.275 \textsc{RLPE} vs.\ 0.230 \texttt{gpt-5.6-sol} and 0.222
\texttt{claude-opus-5}). On both benchmarks, \textsc{RLPE} also beats the prompt
optimizers, even though they select their strongest prompt on the
validation set.
On PlanBench, the trained advisor is competitive with \texttt{gpt-5.6-sol} and both prompt optimizers (0.466 \textsc{RLPE} vs.\
0.435 \texttt{gpt-5.6-sol}, 0.508 GEPA, and 0.515 TextGrad), though
\texttt{claude-opus-5} remains clearly ahead (0.581).

\subsection{Iterative Setting}
\label{sec:results-iteration}

\paragraph{Setup.}
While the from-scratch setting has the advisor write a prompt from
traces alone, the iterative setting also gives it a starting prompt to
revise and the traces produced by running that prompt. We evaluate this
capability with two experiments.

The first experiment gives every advisor model the same starting
prompt, so that each model's iteration ability can be compared
directly. We use the untrained \texttt{qwen-3.5-9b} policy's ten
from-scratch prompts from Section~\ref{sec:results-scratch} as
starting prompts. We choose these prompts because they are genuine
attempts at improvement, so gains over them are not trivial. Still,
the from-scratch results show more effective prompts exist, so
headroom remains for further improvement. Each starting prompt is run
on the same questions as the from-scratch experiment, producing fresh
traces of the starting prompt's performance. Each advisor model then
produces a revised prompt, and we report the mean gain over the
starting prompt's score across the ten batches. We run this
experiment on Enigmata and BBEH only, since one of the untrained
policy's ten PlanBench generations in the from-scratch experiment
failed to parse, leaving no valid starting prompt for that batch.

In the second experiment, each advisor model revises its own
prompts. This tests whether gains can compound when a model is
applied to its own output. We take a model's ten from-scratch
prompts, rerun them to collect
fresh traces, and give each prompt and its traces back to the same model
for a revised prompt. We report the mean test-set scores of the ten
revisions.

\begin{figure}[t]
  \centering
  \includegraphics[width=\textwidth]{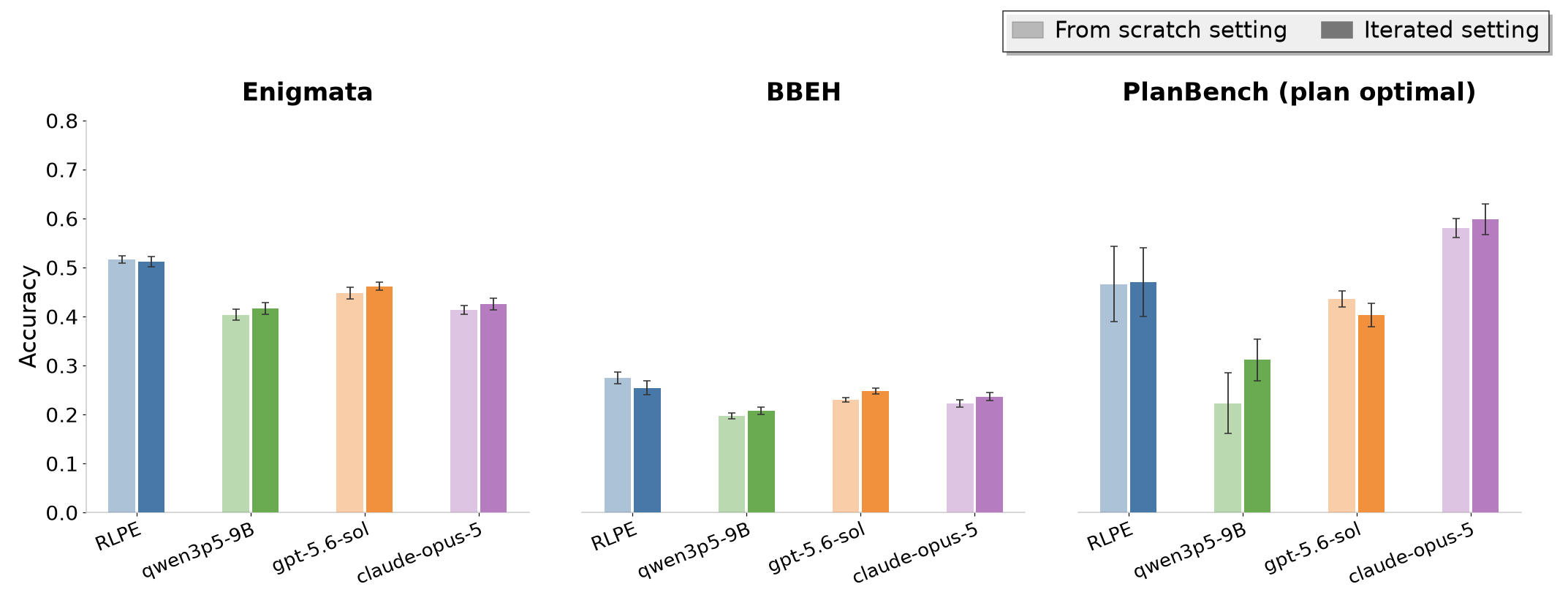}
  \caption{\textbf{Self-iteration results in limited gains.} Each
  advisor model revises its own ten from-scratch prompts. Light bars =
  starting prompts' mean score (from
  Figure~\ref{fig:main-results}); dark bars = revisions' mean
  score, with SE bars. No advisor model gains more than
  3 points on any benchmark, except the untrained policy on
  PlanBench (+9), whose weak starting prompts leave more headroom.}
  \label{fig:self-iteration}
\end{figure}

\paragraph{Results.}
In the first experiment, where every advisor model revises the same starting prompt, the trained advisor shows the largest performance gains
(Figure~\ref{fig:fixed-start}). On Enigmata the trained advisor more
than triples every other model's gain (+9.6 \textsc{RLPE} vs.\ +2.8
\texttt{gpt-5.6-sol} and +2.0 \texttt{claude-opus-5}), while on BBEH it
is competitive with the frontier models (+4.7 \textsc{RLPE} vs.\ +4.4
\texttt{gpt-5.6-sol} and +4.2 \texttt{claude-opus-5}) but still much
better than the untrained policy (+1.3). In the second experiment, models gain little from
revising their own prompts. The gains for almost all
advisor models are under 3 points on every benchmark
(Figure~\ref{fig:self-iteration}). The exception is the untrained
policy on PlanBench, which improves its weak starting prompts by
roughly 9 points.

\subsection{Generalization}
\label{sec:results-generalization}

\subsubsection{Task-Model Generalization}

\paragraph{Setup.}
We test whether the trained advisor still improves task models it never
observed during training. We run each new task model
(\texttt{gpt-oss-120b}, \texttt{gpt-5-nano}, \texttt{gpt-5.4-nano},
\texttt{claude-haiku-4-5}, and \texttt{gemma-4-26b}) on Enigmata. For
each new task model, we collect its traces under the baseline prompt
on the same ten validation batches used in
Section~\ref{sec:results-scratch}. The Enigmata-trained advisor then
writes a from-scratch prompt per batch, and each prompt is scored with
the same task model on the test set. Along with the mean score of the ten
prompts, we report the best of the ten (Best@10) to show whether the
advisor was able to improve the task models at all.

\begin{figure}[t]
  \centering
  \includegraphics[width=0.8\textwidth]{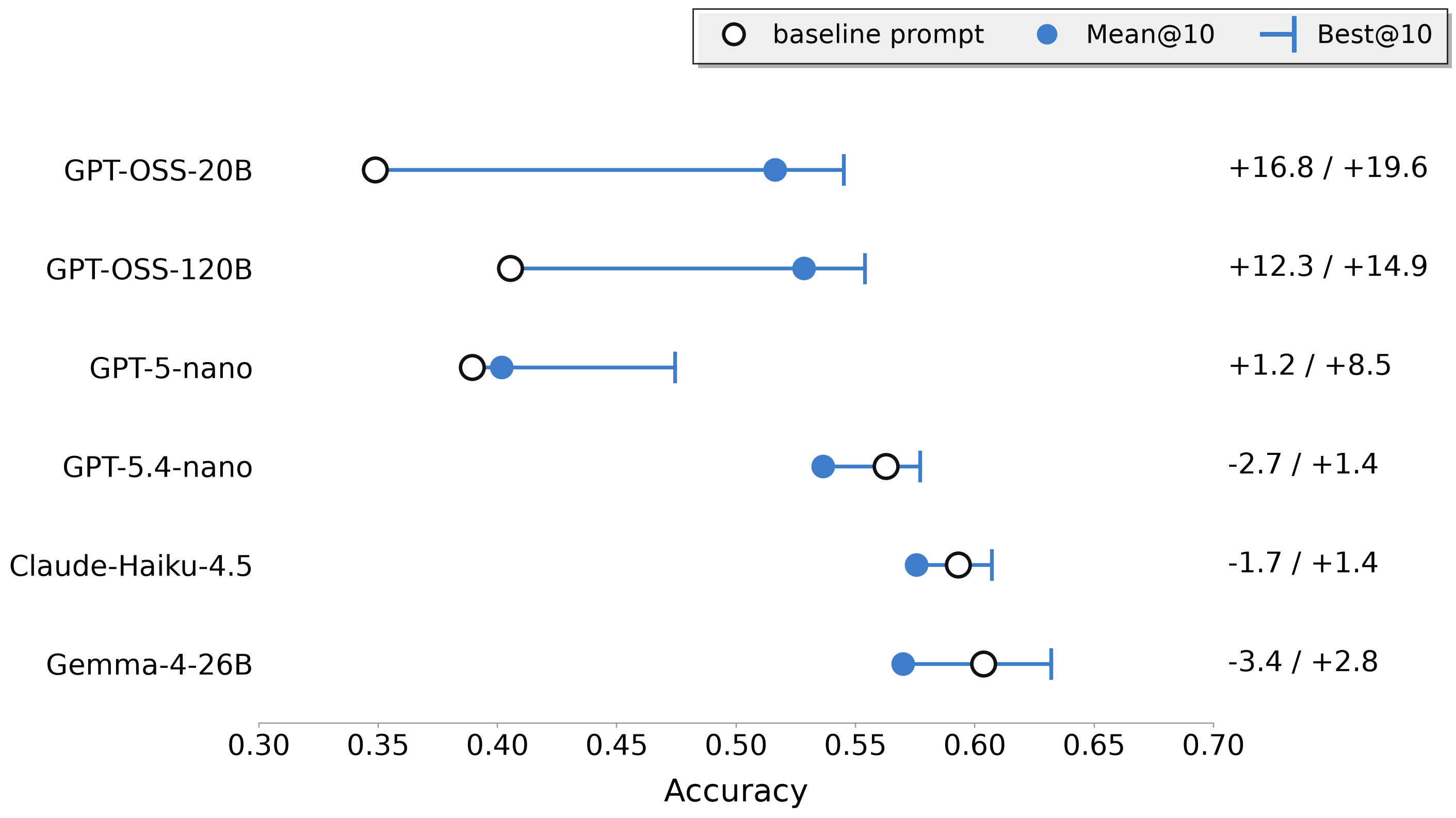}
  \caption{\textbf{The trained advisor improves task models it never
  trained on.} Each row runs the \textsc{RLPE} advisor, trained only
  with \texttt{gpt-oss-20b}, against a new task model on Enigmata. Open circle = baseline-prompt score; filled circle = mean of the
  advisor's ten from-scratch prompts; tick = Best@10;
  right-hand labels = mean / best gains. Mean gains are large on \texttt{gpt-oss-120b}
  (+12.3) despite the model having 6x more parameters than the training task model. The advisor also improves models from
  entirely different families, such as \texttt{gpt-5-nano} (+8.5
  Best@10), though on average it slightly hurts two strong
  task models ($-2.7$ \texttt{gpt-5.4-nano} and $-1.7$
  \texttt{claude-haiku-4-5}).}
  \label{fig:taskmodel-gen}
\end{figure}

\paragraph{Results.}
Figure~\ref{fig:taskmodel-gen} shows the advisor's prompt-writing
capability transfers strongly within the \texttt{gpt-oss} family and
more weakly beyond it.
Despite having six times more total parameters than the training task model,
\texttt{gpt-oss-120b} gains +12.3 points on average, near the
+16.8 on \texttt{gpt-oss-20b}. Gains on other model families occur but are less reliable: the
advisor can achieve a large gain on \texttt{gpt-5-nano} (+8.5
Best@10) but smaller average gains (+1.2 mean). The two
remaining frontier models see small average losses ($-2.7$
\texttt{gpt-5.4-nano} and $-1.7$ \texttt{claude-haiku-4-5}). Similarly,
\texttt{gemma-4-26b} sees a small average loss ($-3.4$), though its
Best@10 remains positive ($+2.8$).

\subsubsection{Dataset Generalization}

\paragraph{Setup.}
We also test whether the learned skill generalizes to a
benchmark never seen during training. We evaluate on JustLogic, a
deductive logic reasoning benchmark, running all three trained advisors from Enigmata, BBEH, and PlanBench under the same from-scratch
protocol, comparing against the other advisor models and prompt optimizers.

\paragraph{Results.}
Despite being trained on entirely different benchmarks, all three
trained advisors land within 1.4 points of one another (0.544
Enigmata-trained, 0.541 PlanBench-trained, and 0.530 BBEH-trained,
Table~\ref{tab:justlogic}) and each beats the untrained policy (0.500). The trained advisors also perform similarly to
\texttt{claude-opus-5} (0.554), though \texttt{gpt-5.6-sol} (0.596)
does better. These results may
suggest training develops a general prompt-writing capability rather
than a domain-specific one, though further study is needed. GEPA and TextGrad also perform well (0.655 and
0.584, respectively).


\section{Discussion and Limitations}
\label{sec:discussion}

\paragraph{Prompt-writing is a learnable skill.} With no custom dataset and relatively low compute,
we trained a 9B-parameter advisor that writes better prompts than far larger frontier models \texttt{gpt-5.6-sol} and
\texttt{claude-opus-5} on two of our three benchmarks. The trained advisor also beats the prompt optimizer tools
on these two benchmarks, even with these tools leveraging frontier model and selecting for the best prompt by evaluating on
the validation set. The trained advisor is also the strongest prompt editor,
improving the same prompt by three times as much
as either frontier model, without ever being trained to revise prompts. These results reflect that writing prompts for another model is a
trainable skill, one that today's frontier models have not
saturated. The
prompting ability of models has received little direct study,
especially for the most recent frontier models, making these results
among the first direct measurements of this capability and pointing
to meaningful room for improvement.

\begin{table}[!t]
  \centering
  \caption{\textbf{Trained advisors generalize to a held-out
  benchmark.} Scores on JustLogic, which no advisor model was trained
  on. Advisor-model rows = mean score of ten from-scratch prompts
  $\pm$ one SE; the three \textsc{RLPE} rows are the trained policies
  from Figure~\ref{fig:main-results}. All three land within 1.4
  points of each other (0.530--0.544), each beating the untrained
  policy (0.500) and staying competitive with \texttt{claude-opus-5}
  (0.554). GEPA (bolded) is the best overall (0.655).}
  \label{tab:justlogic}
  \begin{tabular}{lc}
    \toprule
    Method & Accuracy \\
    \midrule
    baseline prompt & 0.484 \\
    \texttt{qwen-3.5-9b} (untrained) & 0.500 $\pm$ 0.009 \\
    \textsc{RLPE} (BBEH) & 0.530 $\pm$ 0.016 \\
    \textsc{RLPE} (PlanBench) & 0.541 $\pm$ 0.010 \\
    \textsc{RLPE} (Enigmata) & 0.544 $\pm$ 0.010 \\
    \texttt{claude-opus-5} & 0.554 $\pm$ 0.012 \\
    TextGrad & 0.584 \\
    \texttt{gpt-5.6-sol} & 0.596 $\pm$ 0.006 \\
    \textbf{GEPA} & \textbf{0.655} \\
    \bottomrule
  \end{tabular}
\end{table}

\paragraph{The learned skill shows signs of generalization.}
Multiple results indicate the advisor learns a general
prompt-writing skill rather than one tied to its specific training
environment. In addition to prompt iteration emerging without direct
training, the three trained policies are all competitive with
\texttt{claude-opus-5} on the held-out JustLogic benchmark while
clearly beating the untrained policy. Beyond datasets, the skill transfers across task models. Despite
training with a single one (\texttt{gpt-oss-20b}), the advisor
improves task models six times larger and models from other model families.
The advisor even shows gains on \texttt{gpt-5-nano}, demonstrating the
approach extends to closed-source models. Still, the skill transfer is not perfect. The prompt optimizer tools
and one of the frontier models did substantially better on the
held-out dataset, and gains on some of the new task models were
limited. A fuller understanding of this generalization will require
further experimentation, such as training a single advisor across
multiple datasets and task models.


\paragraph{Limitations.}
As one of the few works on training models to write prompts, we
deliberately kept each training run minimal, training on a single
prompting setting, a single benchmark, and a single task model. This
let us isolate the effects of training and study its generalization. We see joint training on both settings (from
scratch and iterative), multi-benchmark training, and training against
multiple task models as natural extensions. Our benchmarks are also
all single-turn tasks with programmatic verifiers. We hope to study
multi-turn and more agentic tasks, including open-ended ones graded by
an LLM judge. Finally, as noted above, our
results are not uniformly strong, with the advisors trailing stronger
baselines on PlanBench and JustLogic. We expect these gaps to narrow with
more training and larger advisor models.

\section{Conclusion}
\label{sec:conclusion}

Using \textsc{RLPE}, we trained a 9B-parameter advisor model that beats
or is competitive with state-of-the-art frontier models and prompt
optimizer tools across multiple benchmarks. The advisor also improves
far larger task models than it trained with, and transfers to a
held-out benchmark. \textsc{RLPE} achieves this by
modeling prompt engineering more faithfully than prior work, rewarding
the advisor for the performance of prompts written from observing the task model's behavior. Our framework is also easy to apply.
Training needs only modest compute and the task dataset itself, since reward is computed directly from performance, and the
advisor writes a prompt from just ten traces, remaining useful when
data is scarce. We see this work as particularly timely
because progress in AI systems increasingly comes from
orchestrating multiple models, and prompting
is a critical part of that orchestration. Our results may suggest this capability remains
underdeveloped in current models, with substantial headroom even for
frontier models. There are many interesting variations on the
\textsc{RLPE} recipe, including adding more prompting
tasks, extending the reward beyond task accuracy to incorporate cost or
latency, and more intelligently sampling the traces the advisor sees.



\clearpage
\bibliographystyle{plainnat}
\bibliography{references}


\clearpage
\appendix

\section{Prompts and Rendering}
\label{app:prompts}

\subsection{Prompt Format Example}
\label{app:prompt-example}

We reproduce the four baseline prompts and one advisor-written example
prompt $p = (s, u)$ below. None of the baselines uses a system prompt.
The BBEH baseline appends one formatting sentence to the question so
that answers can be graded; the Enigmata baseline is the bare question,
since every Enigmata question already carries its own output-format
instructions. The PlanBench baseline is the benchmark's official one-shot query. Its
rules text is fixed across all test instances and shown verbatim; the
worked \texttt{[STATEMENT]}/\texttt{[PLAN]} example varies per instance
(the box shows one instance's), and \texttt{\{\{QUESTION\}\}} is the
instance's own \texttt{[STATEMENT]} block, which ends with an open
\texttt{[PLAN]} tag for the model to complete. On PlanBench, the rules
and one-shot example are shown to the advisor model alongside the
traces during training, and it has to learn whether to include them in
the prompt it generates. The
JustLogic baseline wraps the question (a paragraph of premises and a
statement) with the task framing and a formatting sentence. The final
box is a real prompt the advisor wrote during a BBEH training rollout;
here it splits its instructions between the system prompt and a
user-prompt template that wraps the question.


\begin{promptbox}{BBEH baseline prompt}
=== SYSTEM PROMPT ===

(none)

=== USER PROMPT ===

{{QUESTION}}

Finish your response with a final line of exactly this form: "The answer is X." where X is your final answer.
\end{promptbox}

\begin{promptbox}{Enigmata baseline prompt}
=== SYSTEM PROMPT ===

(none)

=== USER PROMPT ===

{{QUESTION}}
\end{promptbox}

\begin{promptbox}{PlanBench baseline prompt}
=== SYSTEM PROMPT ===

(none)

=== USER PROMPT ===

I am playing with a set of blocks where I need to arrange the blocks into stacks. Here are the actions I can do

Pick up a block
Unstack a block from on top of another block
Put down a block
Stack a block on top of another block

I have the following restrictions on my actions:
I can only pick up or unstack one block at a time.
I can only pick up or unstack a block if my hand is empty.
I can only pick up a block if the block is on the table and the block is clear. A block is clear if the block has no other blocks on top of it and if the block is not picked up.
I can only unstack a block from on top of another block if the block I am unstacking was really on top of the other block.
I can only unstack a block from on top of another block if the block I am unstacking is clear.
Once I pick up or unstack a block, I am holding the block.
I can only put down a block that I am holding.
I can only stack a block on top of another block if I am holding the block being stacked.
I can only stack a block on top of another block if the block onto which I am stacking the block is clear.
Once I put down or stack a block, my hand becomes empty.
Once you stack a block on top of a second block, the second block is no longer clear.

[STATEMENT]
As initial conditions I have that, the red block is clear, the blue block is clear, the yellow block is clear, the hand is empty, the blue block is on top of the orange block, the red block is on the table, the orange block is on the table and the yellow block is on the table.
My goal is to have that the orange block is on top of the blue block.

My plan is as follows:

[PLAN]
unstack the blue block from on top of the orange block
put down the blue block
pick up the orange block
stack the orange block on top of the blue block
[PLAN END]

{{QUESTION}}
\end{promptbox}

\begin{promptbox}{JustLogic baseline prompt}
=== SYSTEM PROMPT ===

(none)

=== USER PROMPT ===

You are given a paragraph of facts/premises, followed by a statement. Using only propositional logic on the paragraph (no prior knowledge; assume all premises are true), determine whether the statement is TRUE (the premises entail it), FALSE (the premises contradict it), or UNCERTAIN (neither).

{{QUESTION}}

End your response with exactly one line: "The answer is TRUE." or "The answer is FALSE." or "The answer is UNCERTAIN."
\end{promptbox}

\begin{promptbox}{Example prompt written by the trained advisor (BBEH, from-scratch)}
=== SYSTEM PROMPT ===

You are an Advanced Reasoning Engine. Your task is to solve complex problems with extreme accuracy. Your behavior must adhere to the following constraints:

1. **NO GUESSING:** You are strictly forbidden from using words or phrases indicating uncertainty. Do not say "I guess," "I think," "probably," "likely," "maybe," "assume," or "it's too complex." If a path leads to a blockage, analyze deeper or revisit previous assumptions, but never stop to output a guess.
2. **FULL COMPUTATION:** For any arithmetic, date, or logic problem requiring calculation or inference, you must trace every step explicitly. Do not skip intermediate steps. If a puzzle has many clues (e.g., >10), list them and verify constraints sequentially.
3. **VERIFICATION:** Before outputting your final answer, mentally simulate the verification step once to ensure it aligns with the derivation.
4. **PERFORMANCE:** You are expected to solve the highest difficulty reasoning tasks. Do not lower your effort based on the apparent complexity of the prompt.
5. **OUTPUT FORMAT:** You will provide a reasoning section followed by a single final line strictly formatted as: "The answer is X." where X is the final calculated or deduced value.

Your primary goal is correctness. Assume every problem is solvable.

=== USER PROMPT ===

You are about to solve a hard reasoning problem. The complexity may be high, but precision is required. 

INSTRUCTIONS:
- Perform all calculations and logical deductions thoroughly.
- If you have to track multiple variables or constraints, create a structured list mentally or via text.
- Do not output any statement that suggests you are stopping early or settling for an estimate.
- Ensure your final answer is derived directly from your reasoning.
- Conclude your response with the exact line: "The answer is X." where X is your result.

---- QUESTION BELOW ----
{{QUESTION}}
\end{promptbox}

\subsection{The From-Scratch Prompt}
\label{app:metaprompt}

The full from-scratch prompt given to the advisor models for each benchmark is reproduced
below. All four share the same skeleton and differ in the task
description, the guidance on reading traces, and the benchmark's
answer-extraction rules. The \texttt{\{\{TRACES\}\}} placeholder is
replaced with the ten rendered traces, and the advisor must emit the
\texttt{\{\{QUESTION\}\}} placeholder verbatim inside the user template
it writes. After the four prompts, we reproduce a single real rendered
trace from an Enigmata baseline run, so the reader can see exactly what
fills \texttt{\{\{TRACES\}\}}; the batch shown to the advisor is ten
such blocks concatenated. In the PlanBench prompt, the puzzle world's
rules and the worked example appear verbatim in the ``WHAT THE MODEL SAW BEFORE EACH QUESTION'' block, since they are necessary for solving the problem. These instructions had to be surfaced somewhere,
and we decided against adding them to each generated prompt automatically. We instead only expose them to the advisor to test whether it learns to include the rules in the prompt it writes, without
which the task model cannot produce plans the grader accepts.


\begin{promptbox}{Enigmata prompt (from-scratch task)}
=== SYSTEM PROMPT ===

You are a master prompt engineer. You write system+user prompt templates for a large language model that solves logic puzzles, and you are judged based on that model's measured accuracy when it uses your prompt.

=== USER MESSAGE ===

Below are 10 real traces from the large language model you are prompting on some logic puzzles. Each trace shows the question exactly as the model saw it, its internal reasoning, its final reply, whether that reply was scored correct or wrong, and the correct answer.

{{TRACES}}

## What the parts of each trace mean

- [QUESTION] is the puzzle exactly as the model received it.
- [REASONING] is the model's internal chain of thought. The grader never sees it. It is where you can watch the model actually think -- whether it had a search strategy or was flailing, whether it mis-transcribed a grid cell by cell, whether it applied a rule backwards, whether it stopped.
- [FINAL OUTPUT] is the visible reply. This is the only thing the verifier reads, and the only thing that gets scored.
- [VERDICT] is whether that reply was scored CORRECT or WRONG by the automatic verifier.
- [GOLD ANSWER] is the correct answer, so you can see exactly where and how the attempt went wrong. Some puzzles have more than one valid solution -- this shows just one correct answer.

Use all of it. Just remember where the points are: only the final output is graded, so a fix that improves the model's thinking but leaves its reply unchanged is worth nothing.

## What you are writing

The same prompt is used verbatim for EVERY task in the benchmark (grid puzzles, logic word problems, ciphers, board placements, arithmetic targets, path finding, ...) by the large language model. Do not specialise it to one puzzle type: it must not hurt the tasks it was not written for.

## How the answer is extracted for the benchmark

An automatic verifier reads the model's reply. It does not read prose:

- For most tasks it takes the last fenced code block in the reply.
- star_battle and campsite are read from <begin_board> ... <end_board> tags.
- zebra_logic is read from the last markdown table in the reply.

Every question already states the exact answer format it requires. Your prompt should help the model comply with the format the question asks for.

## Output format for the two parts of the prompt

Reply with exactly these two tags:

<system_prompt>
...the system prompt for the large language model, or leave empty if you want none...
</system_prompt>

<user_prompt>
...the user prompt template. It MUST contain the literal token {{QUESTION}} exactly where the puzzle text should be substituted...
</user_prompt>

Hard requirements:
1. <user_prompt> MUST contain the literal token {{QUESTION}}. Substitution is a plain string replace, so write it literally -- do not rename it, do not wrap it in quotes or braces of your own.
2. Use these tags, NOT JSON. Prompt text is full of newlines, braces and backticks, and JSON escaping breaks constantly. Raw text inside the tags is shipped exactly as written.
3. Anything outside the two tags is ignored. Put commentary before the first tag if you want it.

Now diagnose what is actually costing accuracy across these 10 traces, then write one system+user prompt pair that fixes as much of it as you can.
\end{promptbox}

\begin{promptbox}{BBEH prompt (from-scratch task)}
=== SYSTEM PROMPT ===

You are a master prompt engineer. You write system+user prompt templates for a large language model that solves hard reasoning problems, and you are judged based on that model's measured accuracy when it uses your prompt.

=== USER MESSAGE ===

Below are 10 real traces from the large language model you are prompting on some hard reasoning problems. Each trace shows the question exactly as the model saw it, its internal reasoning, its final reply, whether that reply was scored correct or wrong, and the correct answer.

{{TRACES}}

## What the parts of each trace mean

- [QUESTION] is the problem exactly as the model received it.
- [REASONING] is the model's internal chain of thought. The grader never sees it. It is where you can watch the model actually think -- whether it worked the problem or gave up, whether it lost track of a long input, whether it applied a rule backwards, whether it stopped.
- [FINAL OUTPUT] is the visible reply. This is the only thing the grader reads, and the only thing that gets scored.
- [VERDICT] is whether that reply was scored CORRECT or WRONG by the automatic grader.
- [GOLD ANSWER] is the correct answer, so you can see exactly where and how the attempt went wrong.

Use all of it. Just remember where the points are: only the final output is graded, so a fix that improves the model's thinking but leaves its reply unchanged is worth nothing.

## What you are writing

The same prompt is used verbatim for EVERY question in the benchmark (multi-hop logic, long buggy tables, boolean expressions, causal judgment, word and spatial puzzles, time arithmetic, humour and sarcasm understanding, ...) by the large language model. Do not specialise it to one task type: it must not hurt the tasks it was not written for.

## How the answer is extracted for the benchmark

An automatic grader reads the model's reply. It does not read prose:

- It looks for a final line of the exact form "The answer is X." and grades only the X.
- If that line is missing, it falls back to reading the FIRST line of the reply, and the answer is usually lost.
- Matching against the correct answer is exact after light normalization. There is no partial credit.

Your prompt must make the model always finish with that exact line.

## Output format for the two parts of the prompt

Reply with exactly these two tags:

<system_prompt>
...the system prompt for the large language model, or leave empty if you want none...
</system_prompt>

<user_prompt>
...the user prompt template. It MUST contain the literal token {{QUESTION}} exactly where the question text should be substituted...
</user_prompt>

Hard requirements:
1. <user_prompt> MUST contain the literal token {{QUESTION}}. Substitution is a plain string replace, so write it literally -- do not rename it, do not wrap it in quotes or braces of your own.
2. Use these tags, NOT JSON. Prompt text is full of newlines, braces and backticks, and JSON escaping breaks constantly. Raw text inside the tags is shipped exactly as written.
3. Anything outside the two tags is ignored. Put commentary before the first tag if you want it.

Now diagnose what is actually costing accuracy across these 10 traces, then write one system+user prompt pair that fixes as much of it as you can.
\end{promptbox}

\begin{promptbox}{PlanBench prompt (from-scratch task)}
=== SYSTEM PROMPT ===

You are a master prompt engineer. You write system+user prompt templates for a large language model that solves planning puzzles, and you are judged based on that model's measured accuracy when it uses your prompt.

=== USER MESSAGE ===

Below are 10 real traces from the large language model you are prompting on some planning puzzles. In every trace, the prompt the model received was: the rules of the puzzle world and one worked example (shown once here, since they were identical in all 10 traces), followed by that trace's question.

===== WHAT THE MODEL SAW BEFORE EACH QUESTION (identical in all 10 traces) =====
I am playing with a set of blocks where I need to arrange the blocks into stacks. Here are the actions I can do

Pick up a block
Unstack a block from on top of another block
Put down a block
Stack a block on top of another block

I have the following restrictions on my actions:
I can only pick up or unstack one block at a time.
I can only pick up or unstack a block if my hand is empty.
I can only pick up a block if the block is on the table and the block is clear. A block is clear if the block has no other blocks on top of it and if the block is not picked up.
I can only unstack a block from on top of another block if the block I am unstacking was really on top of the other block.
I can only unstack a block from on top of another block if the block I am unstacking is clear.
Once I pick up or unstack a block, I am holding the block.
I can only put down a block that I am holding.
I can only stack a block on top of another block if I am holding the block being stacked.
I can only stack a block on top of another block if the block onto which I am stacking the block is clear.
Once I put down or stack a block, my hand becomes empty.
Once you stack a block on top of a second block, the second block is no longer clear.

[STATEMENT]
As initial conditions I have that, the blue block is clear, the orange block is clear, the white block is clear, the hand is empty, the orange block is on top of the yellow block, the yellow block is on top of the red block, the red block is on the table, the blue block is on the table and the white block is on the table.
My goal is to have that the blue block is on top of the white block and the orange block is on top of the red block.

My plan is as follows:

[PLAN]
unstack the orange block from on top of the yellow block
put down the orange block
pick up the blue block
stack the blue block on top of the white block
unstack the yellow block from on top of the red block
put down the yellow block
pick up the orange block
stack the orange block on top of the red block
[PLAN END]
===== END =====

Each trace below shows the question, the model's internal reasoning, its final reply, whether that reply was scored correct or wrong, and a correct plan.

{{TRACES}}

## What the parts of each trace mean

- [QUESTION] is the puzzle statement exactly as the model received it, appended after the shared context above.
- [REASONING] is the model's internal chain of thought. The grader never sees it. It is where you can watch the model actually think -- whether it tracked the state of the stacks or hallucinated it, whether it understood what was being asked, whether it stopped.
- [FINAL OUTPUT] is the visible reply. This is the only thing the grader reads, and the only thing that gets scored.
- [VERDICT] is whether that reply was scored CORRECT or WRONG by the automatic grader.
- [GOLD PLAN] is a correct plan, so you can see exactly where and how the attempt went wrong. Any valid plan is accepted -- this shows just one correct plan (a shortest one).

Use all of it. Just remember where the points are: only the final output is graded, so a fix that improves the model's thinking but leaves its reply unchanged is worth nothing.

## What you are writing

The same prompt is used verbatim for EVERY question in the benchmark by the large language model. Do not specialise it to any single puzzle: it must not hurt the questions it was not written for.

Your prompt is EVERYTHING the model receives besides the question itself -- nothing is added for you. In particular, the rules of the puzzle world only reach the model if your prompt includes them, and without some version of the rules the puzzles are basically impossible to solve. You will likely want to include a version of the rules (it does not have to be the exact official text -- if you can write a version that works better, do). Feel free to also keep the worked example if that helps, but it is not required.

## How the answer is extracted for the benchmark

An automatic grader reads the model's reply and simulates the plan it finds there. It does not read prose:

- It scans EVERY line of the reply. A line becomes one plan step only if it contains exactly one action phrase with fully-named objects, e.g. "unstack the yellow block from on top of the orange block" or "pick up the red block". Argument order is the word order in the line.
- Lines with no action phrase are ignored. A line with several actions in it is dropped entirely.
- There is no way to mark text as "not my plan": any action-phrased line anywhere in the reply -- including inside visible working -- is taken as a plan step, in order.
- The extracted steps are replayed from the initial state; the reply is correct only if every step is legal and the goal holds at the end. Any valid plan counts; it does not have to match the gold plan or be shortest.

## Output format for the two parts of the prompt

Reply with exactly these two tags:

<system_prompt>
...the system prompt for the large language model, or leave empty if you want none...
</system_prompt>

<user_prompt>
...the user prompt template. It MUST contain the literal token {{QUESTION}} exactly where the question text should be substituted...
</user_prompt>

Hard requirements:
1. <user_prompt> MUST contain the literal token {{QUESTION}}. Substitution is a plain string replace, so write it literally -- do not rename it, do not wrap it in quotes or braces of your own.
2. Use these tags, NOT JSON. Prompt text is full of newlines and braces, and JSON escaping breaks constantly. Raw text inside the tags is shipped exactly as written.
3. Anything outside the two tags is ignored. Put commentary before the first tag if you want it.

Now diagnose what is actually costing accuracy across these 10 traces, then write one system+user prompt pair that fixes as much of it as you can.
\end{promptbox}

\begin{promptbox}{JustLogic prompt (from-scratch task)}
=== SYSTEM PROMPT ===

You are a master prompt engineer. You write system+user prompt templates for a large language model that solves logical deduction problems, and you are judged based on that model's measured accuracy when it uses your prompt.

=== USER MESSAGE ===

Below are 10 real traces from the large language model you are prompting on some logical deduction problems. Each trace shows the question exactly as the model saw it, its internal reasoning, its final reply, whether that reply was scored correct or wrong, and the correct answer.

{{TRACES}}

## What the parts of each trace mean

- [QUESTION] is the problem exactly as the model received it.
- [REASONING] is the model's internal chain of thought. The grader never sees it. It is where you can watch the model actually think -- whether it formalized the premises or skimmed them, whether it attempted the derivation or just scanned for a direct mention, whether it treated two similar-sounding claims as the same proposition, whether it stopped.
- [FINAL OUTPUT] is the visible reply. This is the only thing the verifier reads, and the only thing that gets scored.
- [VERDICT] is whether that reply was scored CORRECT or WRONG by the automatic verifier.
- [GOLD ANSWER] is the correct answer, so you can see exactly where and how the attempt went wrong.

Use all of it. Just remember where the points are: only the final output is graded, so a fix that improves the model's thinking but leaves its reply unchanged is worth nothing.

## What you are writing

The same prompt is used verbatim for EVERY problem in the benchmark by the large language model. Every problem has the same shape: a paragraph of natural-language premises, followed by a single statement to judge. The correct answer is exactly one of TRUE (the premises entail the statement), FALSE (the premises contradict the statement), or UNCERTAIN (neither). Problems range from single-step inferences to deeply nested multi-step derivations, and the premises are deliberately worded in varied, sometimes convoluted natural language. The problem must be solved from the paragraph alone -- outside knowledge about whether the claims happen to be true in the real world is irrelevant and misleading. Do not specialise your prompt to one difficulty level or phrasing style: it must not hurt the problems it was not written for.

## How the answer is extracted for the benchmark

An automatic verifier reads the model's reply. It does not read prose:

- It looks for the pattern "The answer is TRUE" / "The answer is FALSE" / "The answer is UNCERTAIN" (case-insensitive) and takes the LAST occurrence in the reply.
- If that pattern is absent anywhere, it falls back to the last bare TRUE / FALSE / UNCERTAIN word in the reply.
- A reply from which neither rule extracts a label is scored wrong.

Your prompt should make the model end its reply with exactly one line of the form "The answer is TRUE." (or FALSE, or UNCERTAIN).

## Output format for the two parts of the prompt

Reply with exactly these two tags:

<system_prompt>
...the system prompt for the large language model, or leave empty if you want none...
</system_prompt>

<user_prompt>
...the user prompt template. It MUST contain the literal token {{QUESTION}} exactly where the problem text should be substituted...
</user_prompt>

Hard requirements:
1. <user_prompt> MUST contain the literal token {{QUESTION}}. Substitution is a plain string replace, so write it literally -- do not rename it, do not wrap it in quotes or braces of your own.
2. Use these tags, NOT JSON. Prompt text is full of newlines, braces and backticks, and JSON escaping breaks constantly. Raw text inside the tags is shipped exactly as written.
3. Anything outside the two tags is ignored. Put commentary before the first tag if you want it.

Now diagnose what is actually costing accuracy across these 10 traces, then write one system+user prompt pair that fixes as much of it as you can.
\end{promptbox}

\begin{promptbox}{Example of a single rendered trace (Enigmata)}
----- TRACE 1 -----
[QUESTION]
Solution for the 24 Game

# Description:
You are given four or five or six integers, ranging from 1 to 13, provide an arithmetic expression that results in 24.

# Rules:

You must use all the given numbers, each exactly once.
The operators you can use include: addition (+), subtraction (-), multiplication (*), and division (/).
You can use parentheses to change the order of operations.

**Response format:**
- Please output your answer within a code block (```) as follows:
```
<result>
```
- If there is a solution, <result> is the sequence of numbers and operators that results in 24, for example: 
```
(8 / 2) * (8 - 2)
```
- If there is no solution, <result> is "cannot form 24".

# Input:
2,2,13,7

Please provide a solution for the 24 game according to the above rules and input.

[REASONING] (internal chain of thought; the verifier never sees this)
We need use 2,2,13,7 to get 24. Try: 13+7=20, 20+2=22, 22+2=24. So (13+7)+(2+2)=24. Output that.

[FINAL OUTPUT] (the visible reply; this is what the verifier reads)
```
(13 + 7) + (2 + 2)
```



[VERDICT]
CORRECT

[GOLD ANSWER]
The answer is: (2+2)+(13+7) = 24
\end{promptbox}

\subsection{The Iteration Prompt}
\label{app:iter-metaprompt}

The iterative task uses the same prompt as the from-scratch task
with small modifications: the prompt being revised is shown in an
additional section, its traces are identified as coming from that
prompt, and the closing instruction asks for a prompt that performs
better than the current one. The full iteration prompts for
Enigmata and BBEH are reproduced below.


\begin{promptbox}{Enigmata prompt (iterative task)}
=== SYSTEM PROMPT ===

You are a master prompt engineer. You write system+user prompt templates for a large language model that solves logic puzzles, and you are judged based on that model's measured accuracy when it uses your prompt.

=== USER MESSAGE ===

Below are 10 real traces from the large language model you are prompting on some logic puzzles. The model was running with the current prompt shown below -- every trace was produced using it. Each trace shows the question exactly as the model saw it, its internal reasoning, its final reply, whether that reply was scored correct or wrong, and the correct answer.

## The current prompt

{{CURRENT_PROMPT}}

## The traces produced with the current prompt

{{TRACES}}

## What the parts of each trace mean

- [QUESTION] is the puzzle exactly as the model received it.
- [REASONING] is the model's internal chain of thought. The grader never sees it. It is where you can watch the model actually think -- whether it had a search strategy or was flailing, whether it mis-transcribed a grid cell by cell, whether it applied a rule backwards, whether it stopped.
- [FINAL OUTPUT] is the visible reply. This is the only thing the verifier reads, and the only thing that gets scored.
- [VERDICT] is whether that reply was scored CORRECT or WRONG by the automatic verifier.
- [GOLD ANSWER] is the correct answer, so you can see exactly where and how the attempt went wrong. Some puzzles have more than one valid solution -- this shows just one correct answer.

Use all of it. Just remember where the points are: only the final output is graded, so a fix that improves the model's thinking but leaves its reply unchanged is worth nothing.

## What you are writing

The same prompt is used verbatim for EVERY task in the benchmark (grid puzzles, logic word problems, ciphers, board placements, arithmetic targets, path finding, ...) by the large language model. Do not specialise it to one puzzle type: it must not hurt the tasks it was not written for.

## How the answer is extracted for the benchmark

An automatic verifier reads the model's reply. It does not read prose:

- For most tasks it takes the last fenced code block in the reply.
- star_battle and campsite are read from <begin_board> ... <end_board> tags.
- zebra_logic is read from the last markdown table in the reply.

Every question already states the exact answer format it requires. Your prompt should help the model comply with the format the question asks for.

## Output format for the two parts of the prompt

Reply with exactly these two tags:

<system_prompt>
...the system prompt for the large language model, or leave empty if you want none...
</system_prompt>

<user_prompt>
...the user prompt template. It MUST contain the literal token {{QUESTION}} exactly where the puzzle text should be substituted...
</user_prompt>

Hard requirements:
1. <user_prompt> MUST contain the literal token {{QUESTION}}. Substitution is a plain string replace, so write it literally -- do not rename it, do not wrap it in quotes or braces of your own.
2. Use these tags, NOT JSON. Prompt text is full of newlines, braces and backticks, and JSON escaping breaks constantly. Raw text inside the tags is shipped exactly as written.
3. Anything outside the two tags is ignored. Put commentary before the first tag if you want it.

Now diagnose what is actually costing accuracy across these 10 traces, then write a new system+user prompt pair that performs even better than the current prompt.
\end{promptbox}

\begin{promptbox}{BBEH prompt (iterative task)}
=== SYSTEM PROMPT ===

You are a master prompt engineer. You write system+user prompt templates for a large language model that solves hard reasoning problems, and you are judged based on that model's measured accuracy when it uses your prompt.

=== USER MESSAGE ===

Below are 10 real traces from the large language model you are prompting on some hard reasoning problems. The model was running with the current prompt shown below -- every trace was produced using it. Each trace shows the question exactly as the model saw it, its internal reasoning, its final reply, whether that reply was scored correct or wrong, and the correct answer.

## The current prompt

{{CURRENT_PROMPT}}

## The traces produced with the current prompt

{{TRACES}}

## What the parts of each trace mean

- [QUESTION] is the problem exactly as the model received it.
- [REASONING] is the model's internal chain of thought. The grader never sees it. It is where you can watch the model actually think -- whether it worked the problem or gave up, whether it lost track of a long input, whether it applied a rule backwards, whether it stopped.
- [FINAL OUTPUT] is the visible reply. This is the only thing the grader reads, and the only thing that gets scored.
- [VERDICT] is whether that reply was scored CORRECT or WRONG by the automatic grader.
- [GOLD ANSWER] is the correct answer, so you can see exactly where and how the attempt went wrong.

Use all of it. Just remember where the points are: only the final output is graded, so a fix that improves the model's thinking but leaves its reply unchanged is worth nothing.

## What you are writing

The same prompt is used verbatim for EVERY question in the benchmark (multi-hop logic, long buggy tables, boolean expressions, causal judgment, word and spatial puzzles, time arithmetic, humour and sarcasm understanding, ...) by the large language model. Do not specialise it to one task type: it must not hurt the tasks it was not written for.

## How the answer is extracted for the benchmark

An automatic grader reads the model's reply. It does not read prose:

- It looks for a final line of the exact form "The answer is X." and grades only the X.
- If that line is missing, it falls back to reading the FIRST line of the reply, and the answer is usually lost.
- Matching against the correct answer is exact after light normalization. There is no partial credit.

Your prompt must make the model always finish with that exact line.

## Output format for the two parts of the prompt

Reply with exactly these two tags:

<system_prompt>
...the system prompt for the large language model, or leave empty if you want none...
</system_prompt>

<user_prompt>
...the user prompt template. It MUST contain the literal token {{QUESTION}} exactly where the question text should be substituted...
</user_prompt>

Hard requirements:
1. <user_prompt> MUST contain the literal token {{QUESTION}}. Substitution is a plain string replace, so write it literally -- do not rename it, do not wrap it in quotes or braces of your own.
2. Use these tags, NOT JSON. Prompt text is full of newlines, braces and backticks, and JSON escaping breaks constantly. Raw text inside the tags is shipped exactly as written.
3. Anything outside the two tags is ignored. Put commentary before the first tag if you want it.

Now diagnose what is actually costing accuracy across these 10 traces, then write a new system+user prompt pair that performs even better than the current prompt.
\end{promptbox}

\section{Data Construction}
\label{app:data}

\subsection{Reasoning-Effort Sensitivity}
\label{app:effort-curve}

\begin{figure}[h]
  \centering
  \includegraphics[width=0.75\textwidth]{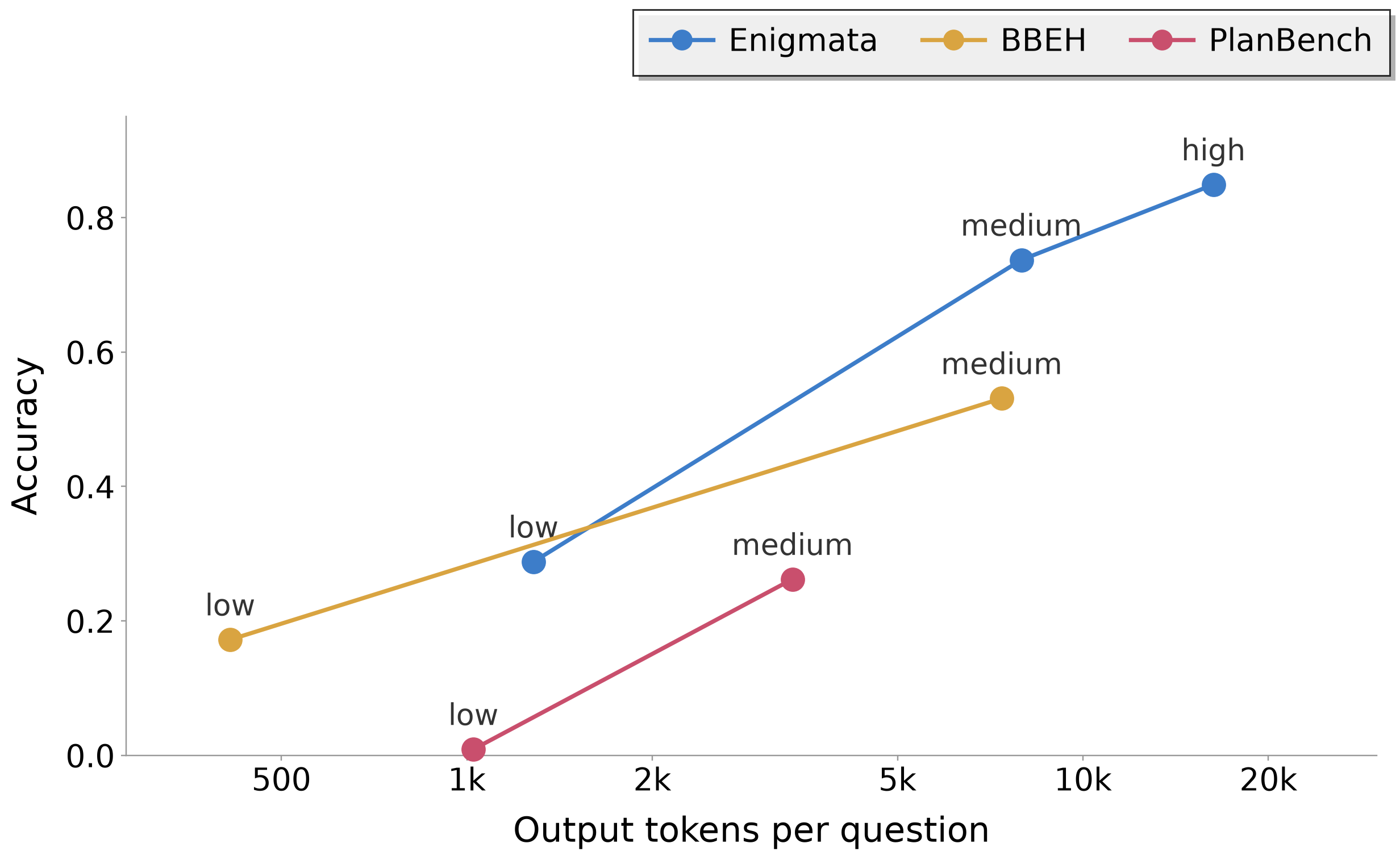}
  \caption{\textbf{Accuracy versus reasoning effort on the training
  benchmarks.} Each point is a \texttt{gpt-oss-20b} run at the given reasoning
  effort under the baseline prompt for each dataset, plotted at its average output
  tokens per question.}
  \label{fig:effort-curve}
\end{figure}

To measure each benchmark's sensitivity to reasoning, we ran
\texttt{gpt-oss-20b} with the baseline prompt over each benchmark's
training data at each reasoning effort, recording accuracy and the
average output tokens per question (Figure~\ref{fig:effort-curve}).
We selected benchmarks where increased reasoning effort produced a
large accuracy gain, roughly 10--15 points or more. At \texttt{high}
reasoning effort the model often reasoned for longer than the maximum
output its context length allows, truncating its responses, so we
omit \texttt{high} measurements for some benchmarks.

\subsection{Split Sizes}
\label{app:splits}

Table~\ref{tab:splits} lists the number of questions in each
benchmark's train, validation, and test sets. The train counts for
Enigmata, BBEH, and PlanBench are after prompt-sensitivity filtering
(Appendix~\ref{app:sensitivity}). We do not train on JustLogic; its
train count is the 100 questions (ten collections of ten) drawn from
its official train split to produce traces for the advisor models, and
it is scored on a 560-question stratified sample of its official
validation split rather than a separate test set. The splits used by
the prompt optimizers are detailed in
Appendix~\ref{app:optimizer-splits}.

\begin{table}[h]
  \centering
  \caption{Train, validation, and test set sizes for each benchmark.}
  \label{tab:splits}
  \begin{tabular}{lrrr}
    \toprule
    Benchmark & Train & Validation & Test \\
    \midrule
    Enigmata  & 1{,}000 & 565 & 565 \\
    BBEH      & 1{,}101 & 678 & 678 \\
    PlanBench & 979     & 601 & 600 \\
    JustLogic & 100     & 560 & --  \\
    \bottomrule
  \end{tabular}
\end{table}

\subsection{Data Splits for the Prompt Optimizers}
\label{app:optimizer-splits}

Table~\ref{tab:optimizer-splits} lists the data GEPA and TextGrad used
on each benchmark. GEPA scores every candidate prompt it accepts during its search on
the selection set and returns the highest-scoring one as its final
prompt. TextGrad instead uses the selection set as a revert gate,
re-scoring the prompt after each update, undoing any update that
lowers the score, and returning the highest-scoring prompt it found. On Enigmata, BBEH, and PlanBench these are the same filtered
train, validation, and test sets used everywhere else in the paper
(Table~\ref{tab:splits}). JustLogic has no separate test set, so its
selection and scoring sets are disjoint partitions of the official
validation split, and its search data is the official train split, the
same one the advisor models' traces are drawn from.

\begin{table}[h]
  \centering
  \caption{Number of questions used by the prompt optimizers on each
  benchmark.}
  \label{tab:optimizer-splits}
  \begin{tabular}{lrrr}
    \toprule
    Benchmark & Search & Selection & Final scoring \\
    \midrule
    Enigmata  & 1{,}000 & 565 & 565 \\
    BBEH      & 1{,}101 & 678 & 678 \\
    PlanBench & 979     & 601 & 600 \\
    JustLogic & 4{,}900 & 490 & 560 \\
    \bottomrule
  \end{tabular}
\end{table}

\subsection{Prompt-Sensitivity Filter}
\label{app:sensitivity}

For each benchmark, we assembled a small set of high-performing prompts
identified during preliminary experimentation, and a small set of
low-performing prompts consisting of the baseline prompt and weaker
prompts from the same experimentation. We then ran five to six runs of
the training questions under each set. For a question $q$, let
$p_{\mathrm{high}}(q)$ and $p_{\mathrm{low}}(q)$ denote the fraction of
high-performing and low-performing runs in which the task model answered
$q$ correctly. A question is kept only if
\[
  p_{\mathrm{high}}(q) > p_{\mathrm{low}}(q),
\]
that is, if the better prompts actually help on it. Questions the task
model always answers correctly or always fails are removed automatically
by this rule, since both fractions tie. The
filter uses training data only; the validation and test sets are left in
each benchmark's natural distribution.

\subsection{PlanBench Optimal-Planning Evaluation}
\label{app:planbench-t2}

PlanBench \citep{valmeekam_planbench_2022} poses block-stacking problems. Each question gives an initial
arrangement of blocks and a goal arrangement, and the model must
produce a sequence of pick-up, put-down, stack, and unstack actions
that reaches the goal. The benchmark defines several official tasks
over the same questions. Its plan-generation task accepts any valid
plan and leaves it ambiguous in the instructions whether a plan longer than necessary is
penalized. The frontier advisor models discovered that it is not penalized. Many
of their prompts instruct the task model to unstack every block onto
the table and then rebuild the goal arrangement from scratch, a
procedure that is valid for any instance but far longer than necessary;
one such prompt states directly that ``extra moves cost nothing -- only
legality and the final goal matter.'' The \texttt{qwen-3.5-9b}-based advisors never
discovered this exploit, and many of the trained advisor's prompts
instead push the task model toward minimal plans.

\begin{figure}[h]
  \centering
  \includegraphics[width=\textwidth]{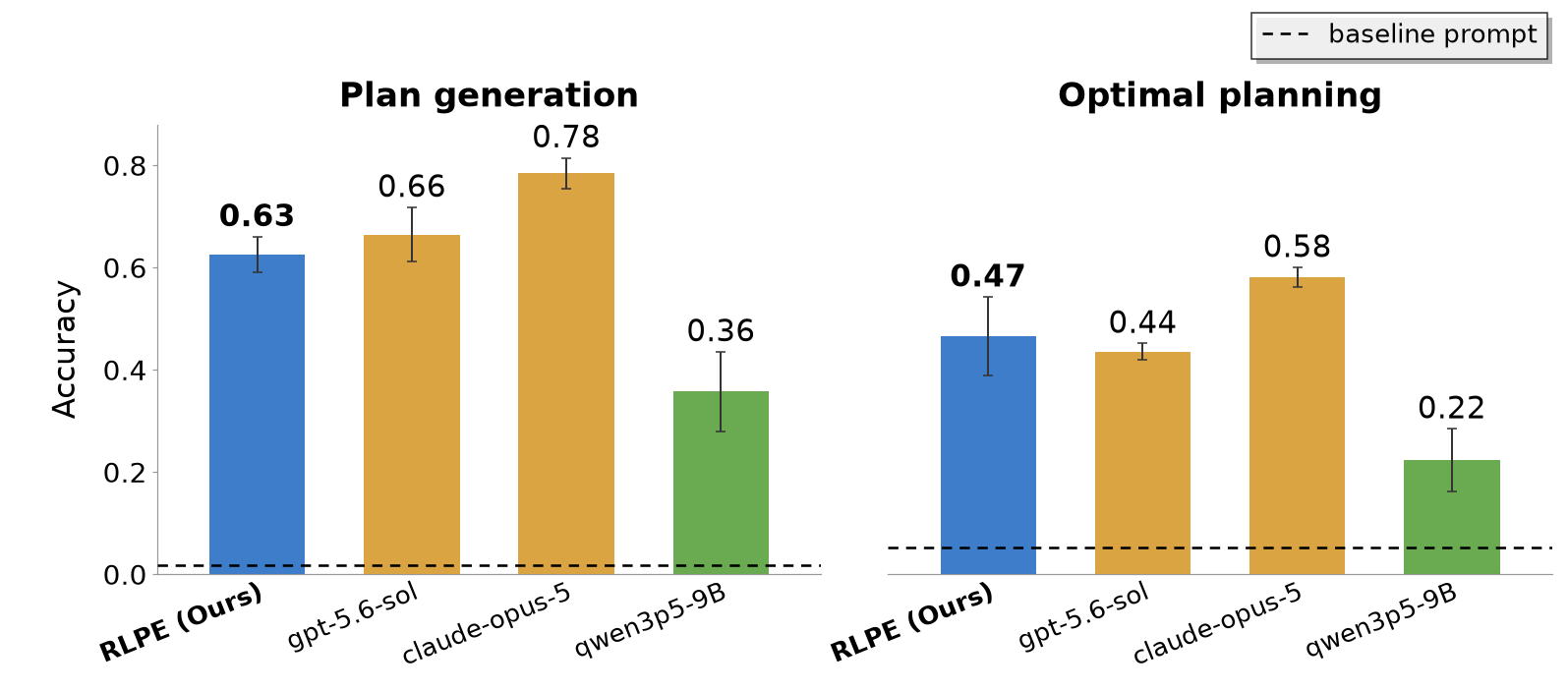}
  \caption{\textbf{Frontier models drop the most when optimality is
  enforced.} PlanBench test accuracy under the benchmark's two
  official tasks, plan generation (any valid plan counts) and optimal
  planning (only a shortest valid plan counts). Bars = mean accuracy
  of each advisor model's ten from-scratch prompts, with SE bars;
  dashed line = baseline prompt performance. Moving to optimal
  planning costs \texttt{gpt-5.6-sol} 23.0 points and
  \texttt{claude-opus-5} 20.3, against 15.9 for \textsc{RLPE}.}
  \label{fig:pb-t1-t2}
\end{figure}

We did not catch this ambiguity until after training, so the advisor
was trained on the plan-generation task. For evaluation we instead use
PlanBench's official optimal-planning task, which poses the same
questions with the ambiguity resolved. Its question text attaches a
one-minute cost to every action in the plan and asks the model to minimize the
total time taken, and its grader accepts a plan only if it is valid and
matches the certified optimal plan length. We view this as the fairer
comparison, since it fixes the ambiguity in the task itself rather than
rewarding whichever advisor happened to guess the grader's behavior.
Figure~\ref{fig:pb-t1-t2} shows the test results under both tasks, with
each advisor model writing its prompts from traces of the corresponding
task. The ordering is largely unchanged, but the frontier models drop
the most once optimality is enforced (\texttt{gpt-5.6-sol} $-23.0$ and
\texttt{claude-opus-5} $-20.3$ points, against $-15.9$ for
\textsc{RLPE}).

\subsection{Compute Details}
\label{app:compute}

All training, and all inference for the advisor and task models, ran
on NVIDIA B200 (180 GB) GPUs on Fireworks AI (cloud); frontier
prompt writers used as advisor models or within a prompt optimizer were accessed through
their native APIs. Each training run used five B200s in total: a
two-GPU trainer that holds and updates the 9B advisor, a single-GPU
deployment serving the current advisor snapshot for rollout sampling,
and two single-GPU deployments serving \texttt{gpt-oss-20b} for
rollout grading. A 50-step run completes in 9--14 hours of wall-clock
time for each of Enigmata, BBEH, and PlanBench. Evaluation used one to
seven additional single-GPU B200 deployments serving the task model,
plus a single-GPU deployment serving each qwen-based prompt writer.

\subsection{Licenses}
\label{app:licenses}

Enigmata's code and evaluation set are released under Apache~2.0, and
its training data under CC~BY-NC~4.0. BBEH's code is released under
Apache~2.0 and its benchmark data under CC~BY~4.0. PlanBench and
JustLogic (code and data) are MIT-licensed. The open-weight models
(\texttt{qwen-3.5-9b}, \texttt{gpt-oss-20b}, \texttt{gpt-oss-120b},
and \texttt{gemma-4-26b}) are all released under Apache~2.0. The
proprietary models (\texttt{gpt-5.6-sol}, \texttt{gpt-5-nano},
\texttt{gpt-5.4-nano}, \texttt{claude-opus-5}, and
\texttt{claude-haiku-4-5}) were accessed through their providers'
public APIs under the corresponding terms of service. GEPA (both the
standalone package and its DSPy implementation) and TextGrad are
MIT-licensed. All assets were used for research purposes in accordance
with these licenses and terms.


\end{document}